%% file: acl.tex
\pdfoutput=1

\documentclass[11pt]{article}

\usepackage[]{acl}

\usepackage{times}
\usepackage{latexsym}

\usepackage[T1]{fontenc}

\usepackage[utf8]{inputenc}

\usepackage{microtype}
\usepackage{amsmath}
\usepackage{amsfonts}
\usepackage{amssymb}
\usepackage{pifont}
\usepackage{graphicx}
\usepackage{booktabs}
\usepackage{multirow}

\newif\ifcomments
\commentstrue
\ifcomments
    \providecommand\ori[1]{[\textcolor{red}{Ori: {#1}}]}
    \providecommand\gal[1]{[\textcolor{brown}{Gal: {#1}}]}
    \providecommand\omer[1]{[\textcolor{purple}{Omer: {#1}}]}
    \providecommand\jb[1]{[\textcolor{blue}{Jonathan: {#1}}]}
    \providecommand\amir[1]{[\textcolor{orange}{Amir: {#1}}]}
\else
    \providecommand{\ori}[1]{}
    \providecommand{\gal}[1]{}
    \providecommand{\omer}[1]{}
    \providecommand{\jb}[1]{}
    \providecommand{\amir}[1]{}
\fi
\newcommand{\ourmodel}{Spider}
\newcommand{\comment}[1]{}
\title{Learning to Retrieve Passages without Supervision}

\author{Ori Ram~~~~Gal Shachaf~~~~Omer Levy~~~~Jonathan Berant~~~~Amir Globerson\\ \\ 
Blavatnik School of Computer Science, Tel Aviv University \\ 
\texttt{ori.ram@cs.tau.ac.il}
}

\begin{document}
\maketitle

\input{00_abstract}
\input{figure_intro}
\input{figure_example}
\input{01_intro}

\input{02_background}

\input{03_method}
\input{04_experiments}

\input{05_results}
\input{06_analysis}
\input{07_related}
\input{08_conclusion}

\section*{Acknowledgements}

We thank Yuval Kirstain and anonymous reviewers for valuable feedback and discussions, and Devendra Singh Sachan for his help with running the ICT and MSS baselines. This project was funded by the European Research Council (ERC) under the European Unions Horizon 2020 research and innovation programme (grant ERC HOLI 819080), the Blavatnik Fund, the Alon Scholarship, the Yandex Initiative for Machine Learning and Intel Corporation.

\bibliography{anthology,custom}
\bibliographystyle{acl_natbib}

\appendix

\input{0A_appendix}



\end{document}

%% file: 00_abstract.tex
\begin{abstract}

Dense retrievers for open-domain question answering (ODQA) have been shown to achieve impressive performance by training on large datasets of question-passage pairs.
In this work we ask whether this dependence on labeled data can be reduced via unsupervised pretraining that 
is geared towards ODQA. We show this is in fact possible, via a novel pretraining scheme designed for retrieval. Our ``recurring span retrieval'' approach uses recurring spans across passages in a document to create pseudo examples for contrastive learning.
Our pretraining scheme directly controls for term overlap across pseudo queries and relevant passages, thus allowing to model both lexical and semantic relations between them.
The resulting model, named \emph{\ourmodel}, performs surprisingly well without any labeled training examples on a wide range of ODQA datasets. 
Specifically, it significantly outperforms all other pretrained baselines in a zero-shot setting, and is competitive with BM25, a strong sparse baseline. 
Moreover, a hybrid retriever over \ourmodel~and BM25 improves over both, and is often competitive with DPR models, which are trained on tens of thousands of examples. Last, notable gains are observed when using \ourmodel~as an initialization for supervised training.\footnote{Our code and models are publicly available:\newline \url{https://github.com/oriram/spider}, and:\newline
\url{https://huggingface.co/tau/spider}}

\end{abstract}

%% file: figure_intro.tex
\begin{figure}[t]
\centering
\hspace*{-20pt}
\includegraphics[width=1.05\columnwidth]{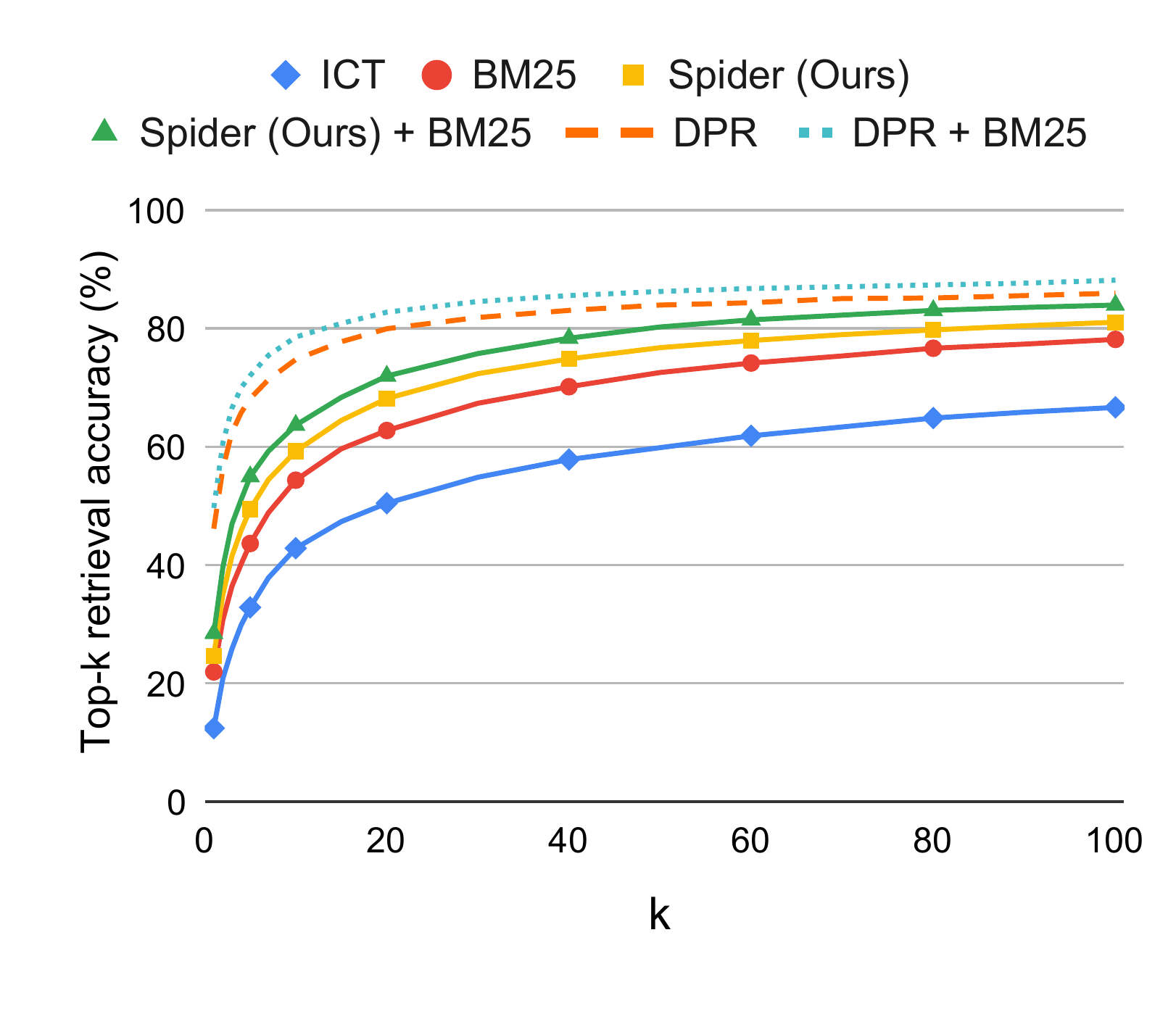}
\vspace{-15pt}
\caption{Top-$k$ retrieval accuracy of various unsupervised methods (solid lines) on the test set of Natural Questions (NQ). DPR (dotted) is supervised (trained on NQ) and given for reference.}
\label{fig:nq}
\end{figure}

%% file: figure_example.tex
\begin{figure*}[t!]
\centering
\includegraphics[width=\textwidth]{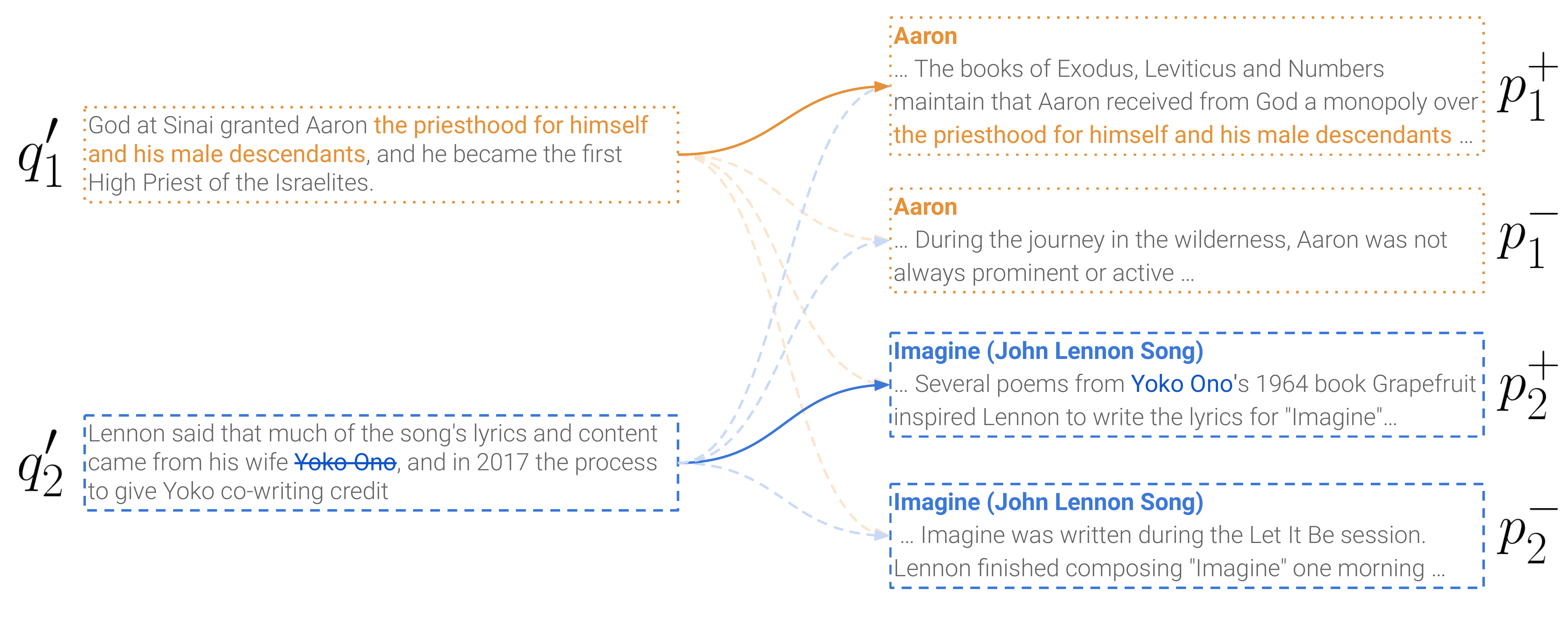}
\caption{An example of our pretraining approach: Given a document $\mathcal{D}$ (e.g. the article \textit{\color{orange}``Aaron''} in Wikipedia), we take two passages that contain a recurring span $S$. One of them is transformed into a short query (left) $q'$ using a random window surrounding $S$, in which $S$ is either kept (top) or removed (bottom). The second passage is then considered the target for retrieval $p^+$, while a random passage from $\mathcal{D}$ that does not contain $S$ is considered the negative $p^-$ (right). Each batch is comprised of multiple such examples, and the pretraining task is to select the passage $p_i^+$ for each query $q'_i$ (solid line) from the passages of all examples (in-batch negatives; dashed lines).  }
\label{fig:example}
\end{figure*}

%% file: 01_intro.tex
\section{Introduction}


State-of-the-art models for retrieval in open domain question answering are based on learning dense text representations  \cite{lee-etal-2019-latent,karpukhin-etal-2020-dense,qu-etal-2021-rocketqa}.
However, such models rely on large datasets of question-passage pairs for training. 
These datasets are expensive and sometimes even impractical to collect (e.g., for new languages or domains), and models trained on them often fail to generalize to new question distributions \cite{sciavolino-etal-2021-simple,reddy2021robust}.

The above difficulty motivates the development of retrieval models that do not rely on large annotated training sets, but are instead trained only on unlabeled text.
Indeed, self-supervision for retrieval has gained considerable attention recently \cite{lee-etal-2019-latent,Guu2020REALMRL,sachan-etal-2021-end,pretraining_ir_survey}. However, when applied in a ``zero-shot'' manner, such models are still outperformed by sparse retrievers like BM25 \cite{robertson2009bm25} and by supervised models (see \citealt{sachan-etal-2021-end}). 
Moreover, models like REALM \cite{Guu2020REALMRL} and MSS \cite{sachan-etal-2021-end,sachan2021emdr} that train a retriever and a reader jointly (i.e. in an end-to-end fashion), treating retrieval as a latent variable, outperform contrastive models like ICT \cite{lee-etal-2019-latent}, but are much more computationally-intensive.

In this work we introduce \emph{\ourmodel} (\textbf{Sp}an-based unsuperv\textbf{i}sed \textbf{de}nse \textbf{r}etriever), a dense model pretrained in a contrastive fashion from \textit{self-supervision only} \cite{contrastive_review}, which achieves retrieval accuracy that significantly improves over unsupervised methods (both contrastive and end-to-end), and is much cheaper to train compared to end-to-end models.

\ourmodel~is based on a novel self-supervised scheme: recurring span retrieval. 
We leverage recurring spans in different passages of the same document (e.g. \textit{\color{blue} ``Yoko Ono''} in Figure~\ref{fig:example}) to create pseudo examples for self-supervised contrastive learning, where one of the passages containing the span is transformed into a short  query that (distantly) resembles a natural question, and the other is the target for retrieval. Additionally, we randomly choose whether to keep or remove the recurring span from the query to explicitly model cases where there is substantial overlap between a question and its target passage, as well as cases where such overlap is small.

We evaluate \ourmodel~on  several ODQA benchmarks.
\ourmodel~narrows the gap between unsupervised dense retrievers and DPR on all benchmarks (Figure~\ref{fig:nq}, Table~\ref{tab:unsupervised}), outperforming all contrastive and end-to-end unsupervised models in top-5 \& top-20 accuracy consistently across datasets. 
Furthermore, we demonstrate that \ourmodel~and BM25 are complementary, and that applying their simple combination \cite{ma2021replication} improves retrieval accuracy over both, sometimes outperforming a supervised DPR model.

We further demonstrate the utility of \ourmodel~as an off-the-shelf retriever via cross-dataset evaluation (i.e., when supervised models are tested against datasets which they were not trained on), a setting that often challenges dense retrievers \cite{sciavolino-etal-2021-simple,reddy2021robust}. In this setting, \ourmodel~is competitive with supervised dense retrievers trained on an abundance of training examples. 

Last, \ourmodel~significantly outperforms other pretrained models when used as an initialization towards DPR training, and also shows strong cross-dataset generalization. For example,  \ourmodel~fine-tuned on TriviaQA is, to the best of our knowledge, the first dense model to outperform BM25 on the challenging EntityQuestions dataset \cite{sciavolino-etal-2021-simple}.

Taken together, our results demonstrate the potential of pretraining for reducing the reliance of ODQA models on training data.

%% file: 02_background.tex
\section{Background}
\label{sec:background}

In open-domain question answering (ODQA), the goal is to find the answer to a given question over a large corpus, e.g. Wikipedia \cite{voorhees-tice-2000-trec,chen-etal-2017-reading,chen-yih-2020-open}. 
This task has gained considerable attention following recent advancement in machine reading comprehension, where models reached human parity in extracting an answer from a paragraph given a question \cite{devlin-etal-2019-bert,raffel-2020-exploring}.

Due to the high cost of applying such reading comprehension models, or \textit{readers}, over the entire corpus, state-of-the-art systems for ODQA first apply an efficient \textit{retriever} -- either sparse \cite{robertson2009bm25,chen-etal-2017-reading} or dense \cite{lee-etal-2019-latent,karpukhin-etal-2020-dense} -- in order to reduce the search space of the reader. 

Recently, dense retrieval models have shown promising results on ODQA, even outperforming strong sparse methods that operate on the lexical-level, e.g. BM25.  Specifically, the dominant approach employs a dual-encoder architecture, where documents and questions are mapped to a shared continuous space such that proximity in that space represents the relevance between pairs of documents and questions. Formally, let $\mathcal{C}=\{p_1,...,p_m\}$ be a corpus of passages. Each passage $p\in \mathcal{C}$ is fed to a passage encoder $E_P$, such that $E_P(p)\in\mathbb{R}^d$. Similarly, the question encoder $E_Q$ is defined such that the representation of a question $q$ is given by $E_Q(q)\in\mathbb{R}^d$. Then, the relevance of a passage $p$ for $q$ is given by:
\begin{equation*}
    s(q,p) = E_Q(q)^\top E_P(p).
\end{equation*}

Given a question $q$, the retriever finds the top-$k$ candidates with respect to $s(q,\cdot)$, i.e. $\text{top-}k_{p\in \mathcal{C}}\  s(q,p)$.
In order to perform this operation efficiently at test time, a maximum-inner product search (MIPS) index \cite{faiss2021} is built over the encoded passages $\{E_P(p_1),...,E_P(p_m)\}$.

While considerable work has been devoted to create pretraining schemes for dense retrieval (\citealt{lee-etal-2019-latent,Guu2020REALMRL}; \textit{inter alia}), it generally assumed  access to large training datasets after pretraining. In contrast, we seek to improve dense retrieval in the challenging unsupervised setting.

Our contribution towards this goal is twofold.
First, we construct a self-supervised pretraining method based on recurring spans across passages in a document to emulate the training process of dual-encoders for dense retrieval. Our pretraining is simpler and  cheaper in terms of compute than end-to-end models like REALM \cite{Guu2020REALMRL} and MSS \cite{sachan-etal-2021-end}.
Second, we demonstrate that a simple combination of BM25 with our models leads to a strong hybrid retriever that rivals the performance of models trained with tens of thousands of examples.

%% file: 03_method.tex
\section{Our Model: \ourmodel}\label{sec:model}

We now describe our approach for pretraining dense retrievers, which is based on a new self-supervised task (Section~\ref{sec:pretraining}). 
Our pretraining is based on the notion of recurring spans \cite{ram-etal-2021-shot} within a document: given two paragraphs with the same recurring span, we construct a query from one of the paragraphs, while the other is taken as the target for retrieval (Figure~\ref{fig:example}). 
Other paragraphs in the document that do not contain the recurring span are used as negative examples.
We train a model from this self-supervision in a contrastive fashion. 

Since sparse lexical methods are known to complement dense retrieval \cite{luan2021sparse,ma2021replication}, we also incorporate a simple hybrid retriever (combining BM25 and \ourmodel) in our experiments (Section~\ref{sec:hybrid}).

\subsection{Pretraining: Recurring Span Retrieval}\label{sec:pretraining}

Given a document $\mathcal{D}\subset\mathcal{C}$ with multiple passages (e.g. an article in Wikipedia), we define \textit{cross-passage recurring spans} in $\mathcal{D}$ as arbitrary n-grams that appear more than once and in more than one passage in $\mathcal{D}$. 
Let $S$ be a cross-passage recurring span in $\mathcal{D}$, and $\mathcal{D}_S\subset\mathcal{D}$ be the set of passages in the document that contain $S$, so $|\mathcal{D}_S| > 1$ by definition. 
First, we randomly choose a \textit{query} passage $q\in\mathcal{D}_S$. In order to resemble a natural language question, we apply a heuristic \textit{query transformation} $T$, which takes a short random window from $q$ surrounding $S$ to get $q'=T(q)$ (described in detail below). 

Similar to DPR, each query has one corresponding positive passage $p^+$ and one corresponding negative passage $p^-$.
For $p^+$, we sample another random passage from $\mathcal{D}$ that contains $S$ (i.e. $p^+\in \mathcal{D}_S \setminus \{q\}$). 
For $p^-$, we choose a passage from $\mathcal{D}$ that does not contain $S$ (i.e. $p^-\in\mathcal{D}\setminus \mathcal{D}_S$). The article title is prepended to both passages (bot not to the query).  

Figure~\ref{fig:example} illustrates this process. 
We focus on the first example (in orange), which is comprised of three passages from the Wikipedia article \textit{\color{orange} ``Aaron''}. The span \textit{\color{orange} ``the priesthood for himself and his male descendants''} appears in two passages in the article.
One of the passages was transformed into a query (denoted by $q'_1$), while the other ($p_1^+$) is taken as a positive passage. Another random passage from the article ($p_1^-$) is considered its negative.

As the example demonstrates, existence of recurring spans in two different passages often implies semantic similarity between their contexts. 

\paragraph{Query Transformation}

As discussed above, after we randomly choose a query passage $q$ (with a recurring span $S$), we apply a query transformation on $q$. The main goal is to make the queries  more ``similar'' to open-domain questions (e.g. in terms of lengths).

First, we define the \textit{context} to keep from $q$. Since passages  are much longer than typical natural questions,\footnote{In our case, passages contain 100 words, while \citet{joshi-etal-2017-triviaqa} report an average length of 14 words for questions.} we take a \textit{random window} containing $S$. The window length $\ell$ is chosen uniformly between 5 and 30 to resemble questions of different lengths. The actual window is then chosen at random from all possible windows of length $\ell$ that contain $S$. 

Second, we randomly choose whether to \textit{keep} $S$ in $q'$ or \textit{remove} it. This choice reflects two complementary skills for retrieval -- the former requires lexical matching (as $S$ appears in both $q'$ and $p^+$), while the latter intuitively encourages semantic contextual representations. 

 The queries in Figure~\ref{fig:example} (left) demonstrate this process. 
 In the top query, the recurring span \textit{\color{orange} ``the priesthood for himself and his male descendants''} was kept as is. 
 In the bottom query, the span \textit{\color{blue} ``Yoko Ono''} was removed.

\paragraph{Span Filtering} To focus on meaningful spans with semantically similar contexts, we apply several filters on recurring spans. First, we adopt the filters from \citet{ram-etal-2021-shot}: (1) spans only include whole words, (2) only maximal spans are considered, (3) spans that contain only stop words are filtered out, (4) spans contain up to 10 tokens. In addition, we add another filter: (5) spans should contain at least 2 tokens. 
Note that in contrast to methods based on \textit{salient spans} \cite{glass-etal-2020-span,Guu2020REALMRL,roberts-etal-2020-much,sachan-etal-2021-end,sachan2021emdr}, our filters do not require a trained model.

\paragraph{Training} At each time step of pretraining, we take a batch of $m$ examples $\{(q'_i,p_i^+,p_i^-)\}_{i=1}^m$, and optimize the cross-entropy loss with respect to the positive passage $p_i^+$ for each query $q'_i$ in a contrastive fashion (i.e., with in-batch negatives), similar to \citet{karpukhin-etal-2020-dense}:
\begin{equation*}
    -\log \frac{\exp\left(s(q'_i,p_i^+)\right)}{\sum_{j=1}^m \left(\exp\big(s(q'_i,p_j^+)\big) + \exp\big(s(q'_i,p_j^-)\big)\right)}
\end{equation*}



\subsection{Hybrid Dense-Sparse Retrieval}\label{sec:hybrid}

It is well established that the strong lexical matching skills of sparse models such as BM25 \cite{robertson2009bm25} are complementary to dense representation models. \citet{ma2021replication} demonstrated strong improvements by using hybrid dense-sparse retrieval, based on BM25 and DPR. Specifically, they define the joint score of a hybrid retriever via a linear combination of the scores given by the two models, i.e.
$s_{\text{hybrid}}(q,p)=s(q,p) + \alpha\cdot \text{BM25}(q,p)$.
They tune $\alpha$ on a validation set of each of the datasets. A similar approach was considered by \citet{luan2021sparse}. Since tuning hyperparameters is unrealistic in our settings, we simply set $\alpha=1.0$ for all hybrid models. Thus, we define:
\begin{equation*}
    s_{\text{hybrid}}(q,p)=s(q,p) + \text{BM25}(q,p)
\end{equation*}

We adopt the normalization technique from \citet{ma2021replication}. We begin by fetching the top-$k'$ (where $k'>k$) passages from each of the models.  
If a passage $p$ is found in the top-$k'$ of a dense retriever but not of BM25, then $\text{BM25}(q,p)$ is set to the minimum value from the top-$k'$ results of BM25 (and vice versa).


%% file: 04_experiments.tex
\section{Experimental Setup}

To evaluate how different retrievers work on different settings and given different amounts of supervision, we simulate various scenarios by using existing datasets, with an emphasis on the unsupervised setting.


\subsection{Datasets}\label{sec:datasets}

We evaluate our method on six datasets commonly used in prior work, all over Wikipedia: Natural Questions (NQ; \citealt{kwiatkowski-etal-2019-natural}), TriviaQA \cite{joshi-etal-2017-triviaqa}, WebQuestions (WQ; \citealt{berant-etal-2013-semantic}), CuratedTREC (TREC; \citealt{curated2015}), SQuAD \cite{rajpurkar-etal-2016-squad} and EntityQuestions (EntityQs; \citealt{sciavolino-etal-2021-simple}). 
The datasets vary significantly in the distribution of questions and the size of training data.

\citet{lewis-etal-2021-question} showed that there exists a significant overlap between train and test questions in ODQA datasets, which poses an issue in our case: supervised models can memorize training questions while unsupervised methods cannot. 
Thus, we also report the results on the ``\textit{no answer overlap}'' portion of the test sets created by \citet{lewis-etal-2021-question} for NQ, TriviaQA and WQ.

\subsection{Baselines}

We consider a variety of baselines, including supervised and self-supervised dense models, as well as sparse methods.
All dense models share the architecture of BERT-base (namely a transformer encoder; \citealt{vaswaniNIPS2017_7181}), including the number of parameters (110M) and uncased vocabulary. In addition, all pretrained dense models use weight sharing between query and passage encoders (only during pretraining). $E_Q(q)$ and $E_P(p)$ are defined as the representation of the \texttt{[CLS]} token. Similar to \citet{gao-callan-2021-condenser}, we do not consider the models trained in \citet{chang2020pretraining}, as they rely on Wikipedia links, and were not made public. 

We now list our baselines (see App.~\ref{app:baselines} for further details). As a sparse baseline model, we follow prior work and take \textit{BM25} \cite{robertson2009bm25}.
We consider several unsupervised dense retrieval models: \textit{ICT} \cite{lee-etal-2019-latent,sachan-etal-2021-end},  \textit{Condenser} and \textit{CoCondenser} \cite{gao-callan-2021-condenser,gao2021unsupervised}. 
We also compare our approach with an unsupervised model trained in an end-to-end fashion (i.e. jointly with a reader): \textit{Masked Salient Spans} (MSS; \citealt{sachan-etal-2021-end,sachan2021emdr}). In addition, we add the results of the unsupervised \textit{Contriever} model \cite{contriever}, a contemporary work.
Last, we add results of \textit{DPR} \cite{karpukhin-etal-2020-dense}, a supervised model, for reference.

\subsection{Evaluation Settings}\label{sec:eval_settings}

We evaluate our method and baselines in a broad range of scenarios. We report top-$k$ retrieval accuracy, i.e. the percentage of questions for which the answer span is found in the top-$k$ passages.

\paragraph{Unsupervised Setting}
Models are trained only on unlabeled data, and evaluated on all datasets without using any labeled examples (i.e. in a zero-shot mode).
As a reference point, we also compare to DPR, which is supervised. 

\paragraph{Cross-Dataset Generalization} To test the robustness of different models across datasets, we compare \ourmodel~to DPR models tested on datasets they were not trained on.\footnote{For unsupervised models, this is essentially equivalent to the unsupervised setting.} The motivation behind these experiments is to determine the quality of all models as ``off-the-shelf'' retrievers, namely on data from unseen distributions of questions. 

\paragraph{Supervised Setting}
We compare \ourmodel~to other pretrained models for retrieval when \textit{fine-tuned} on different amounts of training examples, similar to \citet{karpukhin-etal-2020-dense}. Specifically, we consider the settings where 128 examples, 1024 examples and full datasets are available. We restrict these experiments to NQ and TriviaQA due to the high cost of running them for all datasets and baselines.

\subsection{Implementation Details}

We base our implementation on the official code of DPR \cite{karpukhin-etal-2020-dense}, which is built on Hugging Face Transformers \cite{wolf-etal-2020-transformers}.

\paragraph{Passage Corpus} We adopt the same corpus and preprocessing as \citet{karpukhin-etal-2020-dense}, namely the English Wikipedia dump from Dec. 20, 2018 (following \citealt{lee-etal-2019-latent}) with blocks of 100 words as retrieval units. Preprocessing \cite{chen-etal-2017-reading} removes semi-structured data (e.g., lists, infoboxes, tables, and disambiguation pages), resulting in roughly 21 million passages. 
This corpus is used for both pretraining and all downstream experiments.

\input{table_unsupervised}

\paragraph{Pretraining}

We train \ourmodel~ for 200,000 steps, using batches of size 1024. 
similar to ICT and Condenser, the model is initialized from the uncased BERT-base model, and weight sharing between the passage and query encoders is applied.
Each pseudo-query has one corresponding positive example and one negative example.\footnote{We perform an ablation on this choice in Section~\ref{sec:ablations}.} 
Overall, the model is expected to predict the positive passage out of a total of 2048 passages.\footnote{In-batch negatives are taken across all GPUs, as suggested in \citet{qu-etal-2021-rocketqa}.} 
The learning rate is warmed up along the first 1\% of the training steps to a maximum value of $2\cdot 10^{-5}$, after which linear decay is applied. 
We use Adam \cite{kingma2014adam} with its default hyperparameters as our optimizer, and apply a dropout rate of 0.1 to all layers. 
We utilize eight 80GB A100 GPUs for pretraining, which takes roughly two days. 
In our ablation study (see Section~\ref{sec:ablations}), we lower the learning rate to $10^{-5}$ and the batch size to 512 in order to fit in eight Quadro RTX 8000 GPUs.\footnote{One ablation does involve a batch size of 1,024, and was trained using A100 GPUs as well.} Each ablation takes two days.

\paragraph{Fine-Tuning} 

For fine-tuning, we use the hyperparameters from \citet{karpukhin-etal-2020-dense}, and do not perform any hyperparameter tuning. Specifically, we train using Adam \cite{kingma2014adam} with bias-corrected moment estimates \cite{zhang2021revisiting}, and a learning rate of $10^{-5}$ with warmup and linear decay.
We use batch size of 128 for 40 epochs with two exceptions. 
First, when fine-tuning DPR-WQ and DPR-TREC, we run for 100 epochs for consistency with the original paper. 
Second, when fine-tuning on 128 examples only, we lower the batch size to 32 and run for 80 epochs.\footnote{This is done to avoid running on all examples in each step, which might lead to overfitting. However, we did not test this hypothesis.}
We use BM25 negatives produced by \citet{karpukhin-etal-2020-dense}, and do not create hard negatives by the model itself \cite{xiong2021approximate}. 

\paragraph{Retrieval} When performing dense retrieval, we apply exact search using FAISS \cite{faiss2021}. 
This is done due to the high memory demand of creating an HNSW index for each experiment \cite{karpukhin-etal-2020-dense}.
For sparse retrieval (i.e. BM25), we utilize the Pyserini library \cite{lin2021pyserini}, built on top of Anserini \cite{anserini2017,anserini2018}. For hybrid retrieval, we set $k'=1000$ similar to \citet{ma2021replication}.

\input{table_cross_dataset}

%% file: table_unsupervised.tex
\begin{table*}[t]
\centering
\small
\begin{tabular}{lccccccccc}
\toprule
\multirow{2.6}{55pt}{\textbf{Model}} & 
\multicolumn{3}{c}{\textbf{NQ}} & \multicolumn{3}{c}{\textbf{TriviaQA}} & \multicolumn{3}{c}{\textbf{WQ}}
\\ \cmidrule(lr){2-4} \cmidrule(lr){5-7} \cmidrule(lr){8-10} 
& \textbf{Top-5} & \textbf{Top-20} & \textbf{Top-100} & \textbf{Top-5} & \textbf{Top-20} & \textbf{Top-100}& \textbf{Top-5} & \textbf{Top-20} & \textbf{Top-100}  \\
\midrule
& \multicolumn{9}{c}{\textit{Supervised Models}}  \\
\midrule
DPR-Single   & 68.3 & 80.1 & 86.1 & 71.2 & 79.7 & 85.1 & 62.8 & 74.3 & 82.2 \\ 
DPR-Multi & 67.1 & 79.5 & 86.1 & 69.8 & 78.9 & 84.8 & 64.0 & 75.1 & 83.0  \\ 
DPR-Single + BM25  & 72.2 & 82.9 & 88.3 & 75.4 & 82.4 & 86.5 & 64.4 & 75.1 & 83.1 \\ 
DPR-Mutli + BM25  & 71.9 & 82.6 & 88.2 & 76.1 & 82.6 & 86.5 & 67.3 & 77.2 & 84.5 \\ 
\midrule
& \multicolumn{9}{c}{\textit{Unsupervised Models}}  \\
\midrule
BM25 & 43.8 & 62.9 & 78.3 & 66.3 & 76.4 & 83.2 & 41.8 & 62.4 & 75.5 \\ 
ICT$^*$   & 32.3 & 50.6 & 66.8 & 40.2 & 57.5 & 73.6 & 25.2 & 43.4 & 65.7 \\
Condenser &  13.0 & 25.5 & 43.4 & ~~4.5 & ~~9.6 & 18.5 & 20.3 & 35.8 & 51.9 \\
CoCondenser  & 28.9 & 46.8 & 63.5 & ~~7.5 & 13.8 & 24.3 & 30.2 & 50.7 & 68.7 \\
MSS$^*$ & 41.7 & 59.8 & 74.9 & 53.3 & 68.2 & 79.4 & 29.0 & 49.2 & 68.4 \\
Contriever$^{**}$  & 47.2 & 67.2 &  81.3 & 59.5 &  74.2 &  83.2 & - & - & - \\
\textbf{\ourmodel}  & 49.6 & 68.3 & 81.2 & 63.6 & 75.8 & 83.5 & 46.8 & 65.9 & 79.7 \\ 
\textbf{\ourmodel~+ BM25}  & \textbf{55.1} & \textbf{72.1} & \textbf{84.1} & \textbf{71.7} & \textbf{80.0} & \textbf{85.5} & \textbf{51.0} & \textbf{69.1} & \textbf{81.1} \\ 
\bottomrule
\end{tabular}
\caption{Top-$k$ retrieval accuracy (i.e., the percentage of questions for which the answer is present in the top-$k$ passages) on the test sets of three datasets for supervised and unsupervised approaches. 
DPR-Single is trained on the corresponding dataset only.
We mark in bold the best unsupervised method for each dataset. Further results are given in Tables~\ref{tab:unsupervised_additional_datasets}\&\ref{tab:unsupervised_no_overlap}. $^*$Results reported in \citet{sachan-etal-2021-end,sachan2021emdr}; $^{**}$Results reported in \citet{contriever}.
}
\label{tab:unsupervised}
\end{table*}

%% file: table_cross_dataset.tex
\begin{table*}[t]
\centering
\small
\begin{tabular}{lrcccccc}
\toprule
\textbf{Model} & \textbf{\# Examples} & \textbf{NQ} & \textbf{TriviaQA} & \textbf{WQ} & \textbf{TREC} & \textbf{SQuAD} & \textbf{EntityQs}  \\
\midrule
DPR-NQ & 58,880  & - & 69.0 & 68.8 & 85.9 & 48.9 & 49.7 \\
DPR-TriviaQA & 60,413 & 67.5 & - & 71.4 & 87.9 & 55.8 & 62.7 \\
DPR-WQ & ~~2,474 &  59.4 & 66.7 & - & 82.0 & 52.3 & 58.3 \\
DPR-TREC & ~~1,125 & 57.9 & 64.0 & 61.7 & - & 49.4 & 46.9 \\
DPR-SQuAD & 70,096 & 47.0 & 60.0 & 56.0 & 77.2 & - & 30.9
 \\
DPR-Multi & 122,892 & - & - & - & - & 52.0 & 56.7 \\
\midrule
BM25 & ~~None & 62.9 & 76.4 & 62.4 & 81.1 & \textbf{71.2} & 71.4 \\
ICT & None & 50.6 & 57.5 & 43.4 & - & 45.1 & - \\
\textbf{\ourmodel} & ~~None & 68.3 & 75.8 & 65.9 & 82.6 & 61.0 & 66.3 \\
\textbf{Spider-NQ} & 58,880  & - & \textbf{77.2} & \textbf{74.2} & 89.9 & 57.7 & 61.9
 \\
\textbf{Spider-TriviaQA} & 60,413 & \textbf{75.5} & - & 73.7 & \textbf{91.2} & 68.1 & \textbf{72.9} \\
\bottomrule
\end{tabular}
\caption{Top-20 retrieval accuracy in a cross-dataset ``zero-shot'' setting, where models are evaluated against datasets not seen during their training.
DPR-$x$ and \ourmodel-$x$ are models trained on the full dataset $x$, initialized from BERT and \ourmodel, respectively. DPR-Multi was trained on NQ, TriviaQA, WQ and TREC.
\# Examples is the number of \textit{labeled} examples used to train the model. 
Top-100 retrieval accuracy results are given in Table~\ref{tab:cross-dataset-top100}.
}
\label{tab:cross-dataset}
\end{table*}

%% file: 05_results.tex
\section{Results}\label{sec:results}

Our experiments show that \ourmodel~significantly improves performance in the challenging unsupervised retrieval setting, even outperforming strong supervised models in many cases.
Thus, it enables the use of such retrievers when no examples are available. 
When used for supervised DPR training, we observe significant improvements over the baselines as well.
We perform ablation studies that demonstrate the importance of our pretraining design choices.

\input{table_supervised}
\input{table_ablations}

\subsection{Unsupervised Setting}\label{sec:results_unsup}

Table~\ref{tab:unsupervised} shows the performance of \ourmodel~(measured by top-$k$ retrieval accuracy) compared to other unsupervised baselines on three datasets, \emph{without additional fine-tuning}. Results for remaining datasets are given in Table~\ref{tab:unsupervised_additional_datasets} and Table~\ref{tab:unsupervised_no_overlap}.
Supervised baselines (i.e. DPR) are given for reference. 
Results demonstrate the effectiveness of \ourmodel~w.r.t. other dense pretrained models, across all datasets. 
For example, the average margin between \ourmodel~and ICT is more than 15 points. 
Moreover, \ourmodel~outperforms DPR-Single on three of the datasets (TREC, SQuAD and EntityQs).
When DPR is better than our model, the gap narrows for higher values of $k$.
In addition, it is evident that \ourmodel~is able to outperform BM25 in some datasets (NQ, WQ and TREC), while the opposite is true for others (TriviaQA, SQuAD and EntityQuestions). 
However, our hybrid retriever is able to combine the merits of each of them into a stronger model, significantly improving over both across all datasets. For example, on TriviaQA, \ourmodel~and BM25 achieve 75.8\% and 76.4\% top-20 retrieval accuracy, respectively.
The hybrid model significantly improves over both models and obtains 80.0\%, better than DPR-Single and DPR-Multi (79.7\% and 78.9\%, respectively). 

Moreover, we observe that \ourmodel~consistently surpasses Contriever, with substantial gains for lower values of $k$.



\subsection{Cross-Dataset Generalization}\label{sec:results_cross}

An important merit of \ourmodel~is the fact that a single model can obtain good results across many datasets, i.e. in a ``zero-shot'' setting. 
Table~\ref{tab:cross-dataset} demonstrates the results of supervised models in these scenarios, where DPR models are tested on datasets they were not trained on.
\ourmodel~outperforms four of the six DPR models (DPR-WQ, DPR-TREC, DPR-SQuAD and DPR-Multi) across all datasets. 
In addition, it significantly outperforms DPR-NQ, which is a widely-used retriever,\footnote{The model was downloaded from Hugging Face model hub 200,000 times during December 2021.} on three datasets out of five. Finally, DPR-TriviaQA outperforms \ourmodel~on three datasets.

When fine-tuning \ourmodel~on NQ and TriviaQA (see Sections~\ref{sec:eval_settings};\ref{sec:results_sup}), the resulting models show strong generalization to other datasets. For example,  Spider-NQ outperforms DPR-NQ (initialized from BERT) by 4-12 points. Similar trends are observed for the models trained on TriviaQA. Specifically, Spider-TriviaQA is able to outperform BM25 on EntityQuestions, that is known to challenge dense retrievers \cite{sciavolino-etal-2021-simple}.

\subsection{Supervised Setting}\label{sec:results_sup}

Table~\ref{tab:supervised} shows the performance when fine-tuning pretrained models on 128 examples, 1024 examples and full datasets from NQ and TriviaQA. \ourmodel~establishes notable gains compared to all other dense baselines on both datasets and for all training data sizes. When only 128 examples are available, \ourmodel~significantly outperforms all other models, with absolute gaps of 3-11\% on both datasets. On TriviaQA, \ourmodel~fine-tuned on 128 examples is able to outperform all other baselines when they are trained on 1024 examples. Similar trends are observed for the 1024-example setting (absolute gaps of 1.7-6.2\%).

Even though \ourmodel~was mainly designed for \textit{unsupervised} settings, it outperforms other pretrained models in the full dataset as well. On both datasets, \ourmodel~obtains the best results, improving over DPR models (initialized from BERT) by 1.9-6.5\%.

\subsection{Ablation Study}\label{sec:ablations}\label{sec:results_ablations}

We perform an ablation study on the query transformation applied on the query passage $q$.
We then test the contribution of the negative passage $p^-$ to the performance of our model. Last, we scale up both the batch size and the number of pretraining steps.

\paragraph{Choice of Query Transformation} 
During pretraining, we apply a query transformation on the query $q$. We sample a \textit{random window} containing the recurring span $S$ and either remove or keep $S$. We now test the effect of these choices on our model. We consider two more options for the \textit{context} taken from $q$: (1) the whole passage, for which we replace $S$ with a \texttt{[MASK]} token (as the context is very long, it makes sense to provide the retriever with a signal on what span is sought in the answer), and (2) a prefix of random length preceding $S$, for which we always remove $S$ from the context (as it is in any case, by definition, in the end of $q'$). The top two rows in Table~\ref{tab:ablations} correspond to these ablations. Indeed, both are inferior to taking a random window surrounding $S$ (one before the last row).

In addition, we test whether alternating between keeping and removing $S$ is indeed better than applying only one of them consistently. The third, fourth and fifth rows of Table~\ref{tab:ablations} verify that our motivation was indeed correct: Alternating between the two is superior to each of them on its own.


\paragraph{Effect of Negative Passages} During pretraining, each query $q'_i$ has one positive passage $p_i^+$ and one negative passage $p_i^-$.
We pretrain a model  without negative passages at all, i.e. the target is to select the positive $p_i^+$, given  the positive passages of all other examples $\{p_j^+\}_{j=1}^m$.
This model corresponds to the row with \# $\text{negatives}=0$ (i.e. the sixth row in Table~\ref{tab:ablations}). As expected, the top-$k$ retrieval accuracy of the model drops significantly (2-6\% for different $k$ values) with respect to the same model with \#~$\text{negatives}=1$ as a result of this choice, which is consistent with \citet{karpukhin-etal-2020-dense}.

\paragraph{Scaling up Batch Size and Training Steps} We scale up the batch size and observe improvements of 0.6-1.2\%. We train our model for longer (200K steps instead of 100K), which leads to additional 1.1-1.8\% improvements (last two rows in Table~\ref{tab:ablations}). 

%% file: table_supervised.tex
\begin{table*}[t]
\centering
\small
\begin{tabular}{lccccccccc}
\toprule
\multirow{2.6}{55pt}{\textbf{Model}} & \multicolumn{4}{c}{\textbf{NQ}} && \multicolumn{4}{c}{\textbf{TriviaQA}}  
\\ \cmidrule(lr){2-5} \cmidrule(lr){7-10}
 & \textbf{Top-1} & \textbf{Top-5} & \textbf{Top-20} & \textbf{Top-100} & & \textbf{Top-1} & \textbf{Top-5} & \textbf{Top-20} & \textbf{Top-100} \\
\midrule
BM25 & 22.1 & 43.8 & 62.9 & 78.3 & ~~ & 46.3 & 66.3 & 76.4 & 83.2 \\
\midrule
& \multicolumn{9}{c}{\emph{128 examples}} \\
\midrule
BERT & 12.7 & 27.3 & 43.5 & 60.6 && 16.7 & 33.4 & 49.4 & 65.4 \\
ICT & 22.8 & 45.5 & 64.1 & 78.3 && 32.7 & 54.5 & 68.9 & 79.5 \\
Condenser & 17.6 & 36.8 & 52.7 & 68.6 && 26.1 & 45.9 & 60.2 & 73.7 \\
CoCondenser & 23.2 & 47.9 & 65.2 & 79.2 && 36.3 & 60.1 & 72.8 & 81.6 \\
\textbf{\ourmodel} & \textbf{31.7} & \textbf{57.7} & \textbf{74.3} & \textbf{84.6} && \textbf{47.5} & \textbf{68.5} & \textbf{78.5} & \textbf{85.1} \\
\midrule
& \multicolumn{9}{c}{\emph{1024 examples}}  \\
\midrule
BERT & 26.6 & 49.6 & 65.3 & 78.1 && 32.6 & 52.7 & 66.1 & 77.9 \\
ICT & 30.4 & 55.8 & 72.4 & 83.4 && 38.8 & 60.0 & 72.8 & 82.3 \\
Condenser & 30.8 & 55.1 & 71.7 & 82.2 && 40.7 & 61.1 & 72.4 & 81.2 \\
CoCondenser & 32.7 & 60.1 & 75.6 & 84.8 && 43.3 & 65.4 & 76.2 & 83.6 \\
\textbf{\ourmodel}  & \textbf{37.0} & \textbf{63.0} & \textbf{77.9} & \textbf{86.5} && \textbf{49.5} & \textbf{69.5} & \textbf{79.3} & \textbf{85.5} \\
\midrule
& \multicolumn{9}{c}{\emph{Full Dataset}}  \\
\midrule
BERT  & 46.3 & 68.3 & 80.1 & 86.1 && 53.7 & 71.2 & 79.7 & 85.1 \\
ICT & 46.4 & 69.6 & 80.9 & 87.6 && 55.1 & 72.3 & 80.4 & 85.8 \\
Condenser & 47.0 & 70.1 & 81.4 & 87.0 && 57.4 & 73.4 & 81.1 & 86.1 \\
CoCondenser~~~~~~~~~~ & 47.8 & 70.1 & 80.9 & 87.5 && 58.7 & 75.0 & 82.2 & 86.5 \\
\textbf{\ourmodel} & \textbf{49.4} & \textbf{72.2} & \textbf{82.4} & \textbf{88.0} && \textbf{60.2} & \textbf{76.1} & \textbf{83.1} & \textbf{87.2} \\ 
\bottomrule
\end{tabular}
\caption{Top-$k$ retrieval accuracy of different pretrained models on the test sets of Natural Questions and TriviaQA, \textbf{after fine-tuning} on various sizes of training data: 128 examples, 1024 examples and the full datasets. All models are fine-tuned using the data produced by \citet{karpukhin-etal-2020-dense}, i.e., BM25-based negative examples. }
\label{tab:supervised}
\end{table*}

%% file: table_ablations.tex
\begin{table*}[t]
\centering
\small
\begin{tabular}{lllllcccc}
\toprule
\multicolumn{2}{c}{\textbf{Query Transformation}} &
& \multirow{2.4}{30pt}{\vskip -0.45em \textbf{Batch Size}} & & \multicolumn{4}{c}{\textbf{NQ (Dev Set)}} \\
\cmidrule{1-2}\cmidrule{6-9}
\multicolumn{1}{c}{\textbf{Context}} &
\multicolumn{1}{c}{\textbf{Recurring Span}} & \textbf{\# Negs} &
 & \textbf{\# Steps} & 
  \textbf{Top-1} & \textbf{Top-5} & \textbf{Top-20} & \textbf{Top-100} \\
\midrule
Whole passage & Replace with a \texttt{[MASK]} & 1 & 512 & 100,000 & 17.5 & 36.8 & 52.7 & 67.3 \\
Prefix & Remove & 1 & 512 & 100,000 & 18.9 & 39.7 & 58.0 & 72.4 \\
Random window  & Remove & 1& 512 & 100,000 & 18.6 & 39.2 & 56.8 & 71.7 \\
Random window  & Keep & 1& 512 & 100,000 & 20.3 & 42.0 & 61.1 & 75.9 \\
Random window & Remove / Keep & 1 & 512 & 100,000 & 21.5 & 44.5 & 62.3 & 76.2 \\
\midrule
Random window & Remove / Keep & 0 & 512 & 100,000 & 16.3 & 38.8 & 58.1 & 74.1 \\
Random window & Remove / Keep & 1 & 1,024 & 100,000 & 22.1 & 45.4 & 63.5 & 77.0 \\
\textbf{Random window} & \textbf{Remove / Keep} & 1 & 1,024 & 200,000 & \textbf{23.4} & \textbf{46.5} & \textbf{65.3} & \textbf{78.2} \\
\bottomrule
\end{tabular}
\caption{Ablation study on the development set of Natural Questions. The top rows of the table describe ablations on the \textit{query transformation}:
 We first determine the \textit{context} to take from the query passage, and then decide what operation will be applied on the \textit{recurring span}. The bottom rows of the table study the contribution of the negative passage $p^-$ (\#~Negs $=0$ stands for no negative examples), as well as scaling up the batch size (i.e. the number of queries at each batch) and the total number of training steps.
The last row corresponds to our model \ourmodel. } 
\label{tab:ablations}
\end{table*}

%% file: 06_analysis.tex


%% file: 07_related.tex
\section{Related Work}

Pretraining for dense retrieval has recently gained considerable attention, following the success of self-supervised models in many tasks \cite{devlin-etal-2019-bert,liu2019roberta,brown2020language}. While most works focus on fine-tuning such retrievers on large datasets after pretraining \cite{lee-etal-2019-latent,chang2020pretraining,Guu2020REALMRL,sachan-etal-2021-end,gao-callan-2021-condenser}, we attempt to bridge the gap between unsupervised dense models and strong sparse (e.g. BM25; \citealt{robertson2009bm25}) or supervised dense baselines (e.g. DPR; \citealt{karpukhin-etal-2020-dense}). 
A concurrent work by \citet{oguz2021domain} presented DPR-PAQ, which shows strong results on NQ after pretraining. However, their approach utilizes PAQ \cite{lewis-etal-2021-paq}, a dataset which was generated using models trained on NQ, and is therefore not unsupervised.

Leveraging recurring spans for self-supervised pretraining has previously been considered for numerous tasks, e.g. coreference resolution and coreferential reasoning \cite{kocijan-etal-2019-wikicrem,varkel-globerson-2020-pre,ye-etal-2020-coreferential} and question answering \cite{ram-etal-2021-shot,bian2021bridging,castel2021optimal}. \citet{glass-etal-2020-span} utilize recurring spans \textit{across documents} to create pseudo-examples for QA.

While we focus in this work on dual-encoder architectures, other architectures for dense retrieval have been introduced recently. 
\citet{luan2021sparse} showed that replacing a single representation with multiple vectors per document enjoys favorable theoretical and empirical properties. \citet{khattab2020colbert} introduced late-interaction models, where contextualized representations of query and document tokens are first computed, and a cheap interaction step that models their fine-grained relevance is then applied. 
Phrase-based retrieval \cite{seo-etal-2018-phrase,seo-etal-2019-real} eliminates the need for a reader during inference, as it directly retrieves the answer span given a query. \citet{lee-etal-2021-learning-dense} demonstrated strong end-to-end ODQA results with this approach, and \citet{lee-etal-2021-phrase} showed that it is also effective for passage retrieval. Our pretraining scheme can be seamlessly used for those architectures as well.




%% file: 08_conclusion.tex

\section{Conclusion}

In this work, we explore learning dense retrievers from unlabeled data. Our results demonstrate that existing models struggle in this setup.
We introduce a new pretraining scheme for dual-encoders that dramatically improves performance, reaching good results without any labeled examples. 
Our results suggest that careful design of a pretraining task is important for learning unsupervised models that are effective retrievers for ODQA.

%% file: 0A_appendix.tex
\input{table_unsup_additional}
\input{table_unsup_no_overlap}

\section{Baselines: Further Details}\label{app:baselines}

\paragraph{BM25} \cite{robertson2009bm25} A sparse bag-of-words model that extends TF-IDF (i.e. reward rare terms that appear in both $q$ and $p$) by accounting for document length and term frequency saturation.

\paragraph{BERT} \cite{devlin-etal-2019-bert} was pretrained on two self-supervised tasks: Masked Language Modeling (MLM) and Next Sentence Prediction (NSP). We evaluate BERT only in the supervised setting, namely as a backbone for fine-tuning, similar to DPR.

\paragraph{ICT} \cite{lee-etal-2019-latent} A dual-encoder model which was pretrained on the Inverse Cloze Task. Given a batch of passages, ICT masks a sentence from each passage, and trains to predict what is the source passage for each sentence. ICT encourages lexical matching by keeping the sentence in the original passage with low probability. Note that unlike our approach, ICT is trained to produce representations to \textit{corrupted} passages. In addition, we encourage lexical matching of individual \textit{terms} in the query, rather than the entire query as ICT.

 \citet{sachan-etal-2021-end} trained their own ICT model, which shows stronger performance than \citet{lee-etal-2019-latent}. The authors shared new results with us, in which TREC and EntityQs are missing. Since their model is not public, for fine-tuning we use the model trained by \citet{lee-etal-2019-latent}.

\paragraph{Condenser \& CoCondenser} \cite{gao-callan-2021-condenser,gao2021unsupervised} Condenser is
an architecture that aims to produce dense sequence-level (i.e. sentences and passages) representations via a variant of the MLM pretraining task. Specifically, to predict a masked token $x_t$, they condition the prediction on two representations: (1) a representation of $x_t$ from an earlier layer in the encoder, and (2) a dense sequence-level representation of the \texttt{[CLS]} token at the last layer of the network. CoCondenser adds a ``corpus-aware'' loss alongside MLM to create better embeddings by sampling two sub-spans from each sequence and train in a contrastive fashion.

\input{table_cross_top100}

\paragraph{MSS} \cite{sachan-etal-2021-end,sachan2021emdr} An unsupervised model in which a dense retriever and a reader are trained jointly end-to-end. 
First, \textit{salient spans} (e.g. entities) are identified using a NER model. Then, some of them are masked. The training objective is to predict these missing spans while using retrieved documents as evidence. Due to the latent nature of the retrieval process in this model, its training is substantially more expensive than contrastive learning. In addition, it requires frequent updates of the encoded evidence corpus.    

\paragraph{Contriever} \cite{contriever} A contemporary work. Contriever is an unsupervised dense model trained in a contrastive fashion, using random cropping to generate two views of a given input.

\paragraph{DPR} \cite{karpukhin-etal-2020-dense} A supervised model for ODQA based on dual-encoders and trained in a contrastive fashion (see Section~\ref{sec:background}). All DPR models considered in the paper are initialized with a BERT-base encoder, and trained on \emph{full datasets}: DPR-Single models are trained on a single dataset, and are also referred to as DPR-$x$, where $x$ is the name of the dataset.
DPR-Multi was trained onNQ, TriviaQA, WQ and TREC. 
For DPR-NQ and DPR-Multi, we use the checkpoints released by the authors.
We re-train the other DPR-Single models (which were not made public) using the same hyper-parameters as \citet{karpukhin-etal-2020-dense}. We do not train a DPR model on EntityQs. The models we trained are consistent with the results of \citet{karpukhin-etal-2020-dense}, except for DPR-SQuAD, where we did not manage to reproduce the original results.

\section{Further Results}\label{sec:appendix}

Table~\ref{tab:unsupervised_additional_datasets} and Table~\ref{tab:unsupervised_no_overlap} show the top-$k$ accuracy for the unsupervised setting (complements Table~\ref{tab:unsupervised}) \textit{for additional datasets}. Table~\ref{tab:cross-dataset-top100} shows the top-100 accuracy for the cross-dataset setting (complements Table~\ref{tab:cross-dataset}). 

\section{Limitations \& Risks} 

We point to several limitations and potential risks of \ourmodel.
First, there is still a gap in performance between supervised and unsupervised models, as can be observed in Table~\ref{tab:unsupervised}. 
Second, self-supervised pretraining is heavier in terms of compute than standard supervised training like DPR. Third, \ourmodel~was trained on data solely from Wikipedia, which might hurt its performance when applied to other domains. 
Last, our model may introduce biases as other pretrained language models, e.g. against under-represented groups.

\section{Dataset Statistics} 

Table~\ref{tab:dataset_stats} shows the number of examples in each of the datasets used in our evaluation suite.

\begin{table}[t!]
\small
\centering
\begin{tabular}{lrr}
\toprule
\textbf{Dataset} & \textbf{Train} & \textbf{Test} \\
\midrule
Natural Questions & 58,880 & 3,610 \\
TriviaQA & 60,413 & 11,313 \\
WebQuestions & 2,474 & 2,032 \\
CuratedTREC & 1,125 & 694 \\
SQuAD & 70,096 & 10,570 \\
EntityQs & - & 22,075 \\
\midrule
\multicolumn{3}{c}{\textit{No (Answer) Overlap Datasets}} \\
\midrule
Natural Questions & - & 1,313 \\
TriviaQA & - & 3,201 \\
WebQuestions & - & 856 \\
\bottomrule
\end{tabular}
\caption{Dataset statistics: number of training and test examples in each dataset.}
\label{tab:dataset_stats}
\end{table}

%% file: table_unsup_additional.tex
\begin{table*}[t]
\centering
\small
\begin{tabular}{lccccccccc}
\toprule
\multirow{2.6}{55pt}{\textbf{Model}} & 
\multicolumn{3}{c}{\textbf{CuratedTREC}} & \multicolumn{3}{c}{\textbf{SQuAD}} & \multicolumn{3}{c}{\textbf{EntityQuestions}}
\\ \cmidrule(lr){2-4} \cmidrule(lr){5-7} \cmidrule(lr){8-10} 
& \textbf{Top-5} & \textbf{Top-20} & \textbf{Top-100} & \textbf{Top-5} & \textbf{Top-20} & \textbf{Top-100}& \textbf{Top-5} & \textbf{Top-20} & \textbf{Top-100}  \\
\midrule
& \multicolumn{9}{c}{\textit{Supervised Models}}  \\
\midrule
DPR-Single   & 66.6 & 81.7 & 89.9 & 40.8 & 58.4 & 74.9 & 38.1 & 49.7 & 63.2 \\ 
DPR-Multi & 80.0 & 89.2 & 93.9 & 35.6 & 52.0 & 67.8 & 44.7 & 56.7 & 70.0  \\ 
DPR-Single + BM25  & 75.8 & 87.0 & 93.8 & 66.9 & 77.9 & 86.3 & 61.1 & 71.7 & 81.3 \\ 
DPR-Mutli + BM25  & 84.7 & 90.3 & 95.4 & 58.5 & 72.1 & 83.0 & 63.2 & 73.3 & 82.6 \\ 
\midrule
& \multicolumn{9}{c}{\textit{Unsupervised Models}}  \\
\midrule
BM25 & 64.6 & 81.1 & 90.3 & 57.5 & 71.2 & 82.0 & 61.0 & 71.4 & 80.0 \\ 
ICT$^*$ & ~~~~- & ~~~~- & ~~~~- & 26.5 & 45.1 & 65.2 & ~~~~- & ~~~~- & ~~~~- \\
Condenser &  ~~9.9 & 20.2 & 34.4 & ~~6.1 & 13.2 & 25.3 & ~~1.0 & ~~2.7 & ~~7.6 \\
CoCondenser  & 11.7 & 22.5 & 39.3 & ~~8.5 & 16.5 & 28.8 & ~~0.5 & ~~1.4 & ~~8.7 \\
MSS$^*$ & ~~~~- & ~~~~- & ~~~~- & 33.9 & 51.3 & 68.4 & ~~~~- & ~~~~- & ~~~~- \\
\textbf{\ourmodel}  & 65.9 & 82.6 & 92.8 & 43.6 & 61.0 & 76.0 & 54.5 & 66.3 & 77.4 \\ 
\textbf{\ourmodel~+ BM25}  & \textbf{74.5} & \textbf{86.5} & \textbf{93.9} & \textbf{60.9} & \textbf{74.6} & \textbf{84.5} & \textbf{65.4} & \textbf{75.0} & \textbf{82.6} \\ 
\bottomrule
\end{tabular}
\caption{Results for an evaluation setup as in Table~\ref{tab:unsupervised} for the remaining datasets. Top-$k$ retrieval accuracy (i.e., the percentage of questions for which the answer is present in the top-$k$ passages) for supervised and unsupervised approaches. 
DPR-Single is trained on the corresponding dataset only.
We mark in bold the best unsupervised method for each dataset. $^*$Results shared with us by the authors of \citet{sachan-etal-2021-end,sachan2021emdr}.
}
\label{tab:unsupervised_additional_datasets}
\end{table*}

%% file: table_unsup_no_overlap.tex
\begin{table*}[t]
\centering
\small
\begin{tabular}{lccccccccc}
\toprule
\multirow{2.6}{55pt}{\textbf{Model}} & 
\multicolumn{3}{c}{\textbf{NQ (No Overlap)}} & \multicolumn{3}{c}{\textbf{TriviaQA (No Overlap)}} & \multicolumn{3}{c}{\textbf{WQ (No Overlap)}}
\\ \cmidrule(lr){2-4} \cmidrule(lr){5-7} \cmidrule(lr){8-10} 
& \textbf{Top-5} & \textbf{Top-20} & \textbf{Top-100} & \textbf{Top-5} & \textbf{Top-20} & \textbf{Top-100}& \textbf{Top-5} & \textbf{Top-20} & \textbf{Top-100}  \\
\midrule
& \multicolumn{9}{c}{\textit{Supervised Models}}  \\
\midrule
DPR-Single   & 54.5 & 68.7 & 76.8 & 48.0 & 56.4 & 62.6 & 47.2 & 60.0 & 71.1 \\ 
DPR-Multi & 54.2 & 68.8 & 77.1 & 46.9 & 55.8 & 62.4 & 48.0 & 60.5 & 70.4  \\ 
DPR-Single + BM25  & 60.9 & 74.0 & 81.1 & 53.4 & 60.2 & 64.9 & 50.5 & 62.4 & 72.8 \\ 
DPR-Mutli + BM25  & 61.7 & 73.7 & 80.9 & 54.6 & 60.9 & 65.2 & 53.3 & 63.8 & 73.0 \\ 
\midrule
& \multicolumn{9}{c}{\textit{Unsupervised Models}}  \\
\midrule
BM25 & 38.8 & 55.5 & 70.1 & 47.3 & 56.0 & 62.4 & 35.4 & 53.6 & 66.4 \\ 
ICT$^*$ & 27.6 & 44.1 & 58.8 & 26.2 & 38.5 & 51.5 & 19.4 & 33.2 & 52.5 \\
Condenser &  ~~6.5 & 13.3 & 26.6 & ~~1.8 & ~~4.2 & ~~9.1 & ~~9.0 & 19.3 & 30.8 \\
CoCondenser  & 23.8 & 36.8 & 52.4 & ~~4.6 & ~~8.0 & 13.8 & 21.3 & 38.3 & 54.2 \\
MSS$^*$ & 33.2 & 49.7 & 66.1 & 36.2 & 47.9 & 58.0 & 19.7 & 36.9 & 54.1 \\
\textbf{\ourmodel}  & 44.3 & 60.7 & 73.7 & 45.2 & 55.6 & 62.7 & 38.1 & 54.8 & 69.7 \\ 
\textbf{\ourmodel~+ BM25}  & \textbf{49.3} & \textbf{65.2} & \textbf{77.6} & \textbf{51.9} & \textbf{59.4} & \textbf{64.6} & \textbf{42.9} & \textbf{58.4} & \textbf{72.0} \\ 
\bottomrule
\end{tabular}
\caption{Top-$k$ retrieval accuracy (i.e., the percentage of questions for which the answer is present in the top-$k$ passages) on the ``\textit{no-answer-overlap}'' portion of the test sets of three datasets \cite{lewis-etal-2021-question} for supervised and unsupervised approaches. 
DPR-Single is trained on the corresponding dataset only.
We mark in bold the best unsupervised method for each dataset. $^*$Results shared with us by the authors of \citet{sachan-etal-2021-end,sachan2021emdr}.
}
\label{tab:unsupervised_no_overlap}
\end{table*}

%% file: table_cross_top100.tex
\begin{table*}[t]
\centering
\small
\begin{tabular}{lrcccccc}
\toprule
\textbf{Model} & \textbf{\# Examples} & \textbf{NQ} & \textbf{TriviaQA} & \textbf{WQ} & \textbf{TREC} & \textbf{SQuAD} & \textbf{EntityQs}  \\
\midrule
DPR-NQ & 58,880  & - & 78.7 & 78.3 & 92.1 & 65.2 & 63.2 \\
DPR-TriviaQA & 60,413 & 79.7 & - & 81.2 & 93.7 & 71.1 & 74.6 \\
DPR-WQ & ~~2,474 &  72.6 & 77.9 & - & 90.8 & 67.6 & 70.2 \\
DPR-TREC & ~~1,125 & 71.0 & 76.0 & 74.6 & - & 65.3 & 61.1 \\
DPR-SQuAD & 70,096 & 65.1 & 75.6 & 72.9 & 89.5 & - & 49.3 \\
DPR-Multi & 122,892 & - & - & - & - & 67.8 & 70.0 \\
\midrule
BM25 & ~~None & 78.3 & 83.2 & 75.5 & 90.3 & \textbf{82.0} & 80.0 \\
ICT & None & 66.8 & 73.6 & 65.7 & - & 65.2 & - \\
\textbf{\ourmodel} & ~~None & 81.2 & 83.5 & 79.7 & 92.8 & 76.0 & 77.4 \\
\textbf{Spider-NQ} & 58,880  & - & \textbf{83.7} & 82.5 & 94.1 & 72.8 & 74.1  \\
\textbf{Spider-TriviaQA} & 60,413 & \textbf{85.0} & - & \textbf{83.3} & \textbf{95.4} & 80.6 & \textbf{81.4} \\
\bottomrule
\end{tabular}
\caption{Results for an evaluation setup as in Table~\ref{tab:cross-dataset}, measured by top-100 retrieval accuracy in a cross-dataset ``zero-shot'' setting, where models are evaluated against datasets not seen during their training.
DPR-$x$ is a model trained on the full dataset $x$, and DPR-Multi was trained on NQ, TriviaQA, WQ and TREC.
\# Examples is the number of \textit{labeled} examples used to train the model.
}
\label{tab:cross-dataset-top100}
\end{table*}